\documentclass{svproc}
\usepackage{url}

\usepackage{hyperref}
\usepackage{amsmath}
\usepackage{color,soul}
\usepackage{booktabs}
\usepackage{caption}
\usepackage{subcaption}
\usepackage{amssymb}
\usepackage{setspace}
\usepackage{mathtools}
\usepackage[numbers]{natbib}

 \newcounter{todocnt}

\begin{document}
\mainmatter              
\title{Predicting Early Dropouts of an Active and Healthy Ageing App}
\titlerunning{Predicting Early Dropouts of an AHA App}  
%
\author{Vasileios Perifanis\inst{1} \and Ioanna Michailidi\inst{1} \and Giorgos Stamatelatos\inst{1} \and\\George Drosatos\inst{2} \and Pavlos S. Efraimidis\inst{1, 2}}
\authorrunning{Perifanis et al.} 
%
\tocauthor{Vasileios Perifanis, Ioanna Michailidi, Giorgos Stamatelatos, George Drosatos, Pavlos S. Efraimidis}

\institute{Department of Electrical and Computer Engineering,\\Democritus University of Thrace, Xanthi, 67100, Greece\\
\email{\{vperifan, ioanmich8, gstamat, pefraimi\}@ee.duth.gr}
\and
Institute for Language and Speech Processing,\\Athena Research Center, Xanthi, 67100, Greece\\
\email{gdrosato@athenarc.gr}
}

\maketitle              

\begin{abstract}
In this work, we present a machine learning approach for predicting early dropouts of an active and healthy ageing app. The presented algorithms have been submitted to the IFMBE Scientific Challenge 2022 (\href{https://wchallenge2022.lst.tfo.upm.es}{wchallenge2022.lst.tfo.upm.es}), part of IUPESM WC 2022. We have processed the given database and generated seven datasets. We used pre-processing techniques to construct classification models that predict the adherence of users using dynamic and static features. We submitted 11 official runs and our results show that machine learning algorithms can provide high-quality adherence predictions. Based on the results, the dynamic features positively influence a model's classification performance. Due to the imbalanced nature of the dataset, we employed oversampling methods such as SMOTE and ADASYN to improve the classification performance. The oversampling approaches led to a remarkable improvement of 10\%. Our methods won first place in the IFMBE Scientific Challenge 2022.
\end{abstract}
\section{Introduction}
The goal of the IFMBE Scientific Challenge 2022\footnote{\url{https://wchallenge2022.lst.tfo.upm.es}} was to capture the patterns of user acquisitions to identify early dropouts. More precisely, given a window of $n=12$ consecutive scheduled acquisitions $S = \{S_1, S_2, \dots, S_n\}$, where $S_i$ denotes the number of received measurements, along with other static features $D$, such as demographic characteristics, the objective is to predict the user's adherence $A$ for the following three future acquisitions $FS = \{FS_1, FS_2, FS_3\}$, where $FS_i \in \{0, 1\}$. On a higher level, the adherence prediction is taken as a binary classification task. The adherence is considered \textit{low} when the target user has one or zero acquisitions in the following three acquisitions and \textit{high}, otherwise. In other words, given $S_t^u$ scheduled acquisitions for a user $u$ at time step $t$, the adherence is calculated as follows:
\begin{equation}
    A_t^u=
    \begin{cases}
      0, & \text{if}\ \sum_{i=1}^{3} FS_{t+i}^{u} < 2 \\
      1, & \text{otherwise}
    \end{cases}
\end{equation}

In this work, we employ machine learning algorithms to identify early dropouts using the dynamic states of acquisitions fused with user-specific characteristics, such as the age and gender as well as user responses to questionnaires.

The rest of this work is structured as follows. Section \ref{methods} describes the methods for dataset pre-processing. Our local evaluation and official results of IFMBE Scientific Challenge 2022 are presented in Section \ref{evaluation}. Finally, Section \ref{conclusion} summarizes and concludes our work.

\section{Pre-processing Methods}
\label{methods}
In this section, we give a brief overview of the dataset and describe the pre-processing methods.

\subsection{Dataset and Features}
The MAHA dataset is described in detail in Fico et al. ~\cite{Fico_2022_MAHA}. It contains $\approx 400$ users with acquisitions, demographic features and answered questionnaires. In summary, the data is organized in 10 tables; four tables contain the users' acquisitions per activity, two with questions concerning the acceptability of users to the application, two with questions concerning the participants' quality of life, one table contains the socio-demographic characteristics and the last table contains the application logs.
Below, we provide a brief overview of the provided questionnaires to demonstrate which features are included later, in the final training datasets of this work. 

The acceptability questionnaires are the Self Perception Questionnaire (SPQ) and the UTAUT, containing 6 and 31 questions, respectively. The SPQ is administered at the beginning and the end of the experiment (instances 1 and 3, respectively). The UTAUT is only administered at the end of the experiment.

The quality of life questionnaires are the EQ5D3L questionnaire and the UCLA, containing 5 and 20 questions, respectively. Both EQ5D3L and UCLA are administered at instances 1 and 3 of the experiment.

The IFMBE Scientific Challenge 2022 comprises two phases with 15 allowed attempts in total (5 for Phase I and 10 for Phase II). In Phase II, we are provided with additional data to improve our classification algorithms. The rest of this work discusses the data using the combined datasets from Phases I and II.

\subsection{Pre-processing}\label{subsec:preprocessing}

The provided database should be pre-processed before using a machine learning algorithm. The goal of pre-processing is to improve the quality of the data, for example by handling null values and normalizing the features.

\paragraph{\textbf{Dynamic Features.}}
First, we processed the dynamic features of users, i.e., the number of acquisitions per activity. The initial step of our data pre-processing approach is data cleansing. Thus, we discarded users with status that differ from the three distinct values clarified in the socio-demographic table (\textit{Still using technology}, \textit{Finished}, or \textit{Dropout}). Moreover, users who did not interact with any of the 4 activities (\textit{Brain-games}, \textit{Finger-tapping}, \textit{Mindfulness}, or \textit{Physical} activity) or who interacted with the MAHA network during a period of fewer than 6 weeks, were not taken into consideration and removed. Following this process, the number of users was reduced to 463. Note that we treated the additional data provided in Phase II as new 'unseen' samples.

The generated acquisitions after the data cleansing had timestamps dating from August 2018 to March 2021. For each user, we utilized the first and last date they had interacted with any of the 4 activities. The dates were then rounded to the previous Monday or Sunday, respectively. The interval between the rounded first and last dates produces the active period with varying lengths for each participant.

Furthermore, we made the following assumption regarding the acquisitions. Given that data acquisitions followed a protocol composed of scheduled acquisitions and the prerequisite that participants should have made at least 2 acquisitions per week, we divided weeks into 2 sessions: Monday to Thursday and Friday to Sunday. 

We divided the active period into such sessions for each user using the assumption mentioned above. Additionally, for each session, we calculated the number of activities that were successfully performed during this section. 

Finally, we produced 15-session sets using all the possible linear combinations in each participant’s active period by applying a sliding window algorithm. The final dataset consists of 84111 rows (session acquisitions). The last three acquisitions were added together to produce the corresponding target adherence.

\begin{figure*}[htb!]
    \centering
    \includegraphics[width=0.6\textwidth]{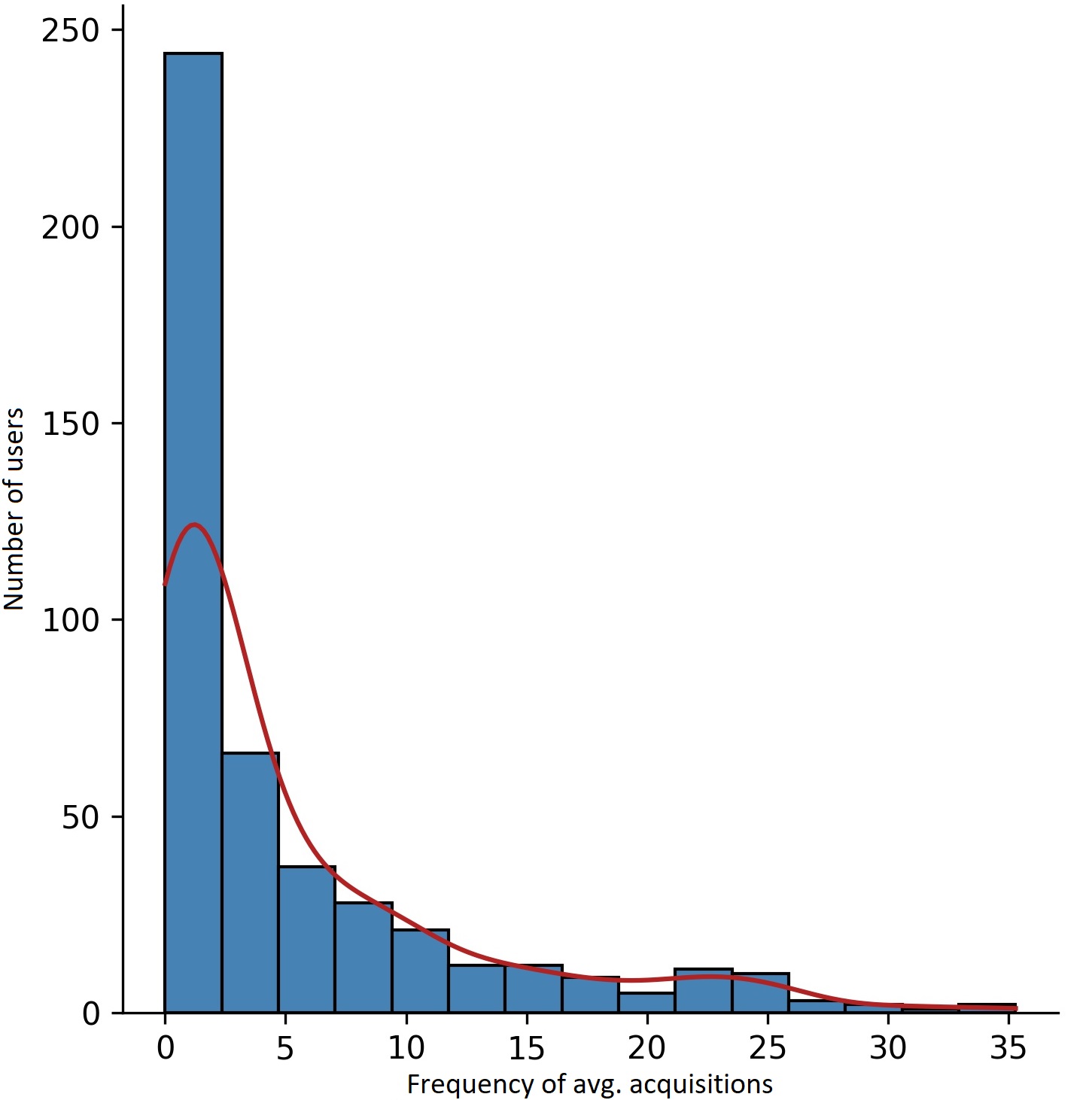}
\caption{Distribution of average number of effective acquisitions per user.}
\label{fig:avg}
\end{figure*}

After collecting the acquisitions, we calculated the average acquisitions per $n=12$ consecutive scheduled acquisitions per user. The average number of acquisitions $\overline{|S|}$ considering all users is 5.24, the minimum average acquisitions $\overline{|S|}_{min}$ is 0.0054 and the maximum average acquisitions $\overline{|S|}_{max}$ is 35.29. Figure~\ref{fig:avg} shows the distribution of average acquisitions per user. Most users (315 out of 463) have less than five average acquisitions per 12-tuple session and 65 users have at least 12, which corresponds to one acquisition per session, i.e., $\approx 4$ per month. Intuitively, users with more than 11 acquisitions per 12-tuple session are users with high adherence and therefore, most users tend to present low application usage rates. Note that the user identifiers were removed in the training stage of machine learning models since the problem is to exploit acquisition patterns and generalize on unseen samples.

\begin{figure*}[htb!]
\centering
\includegraphics[width=\textwidth]{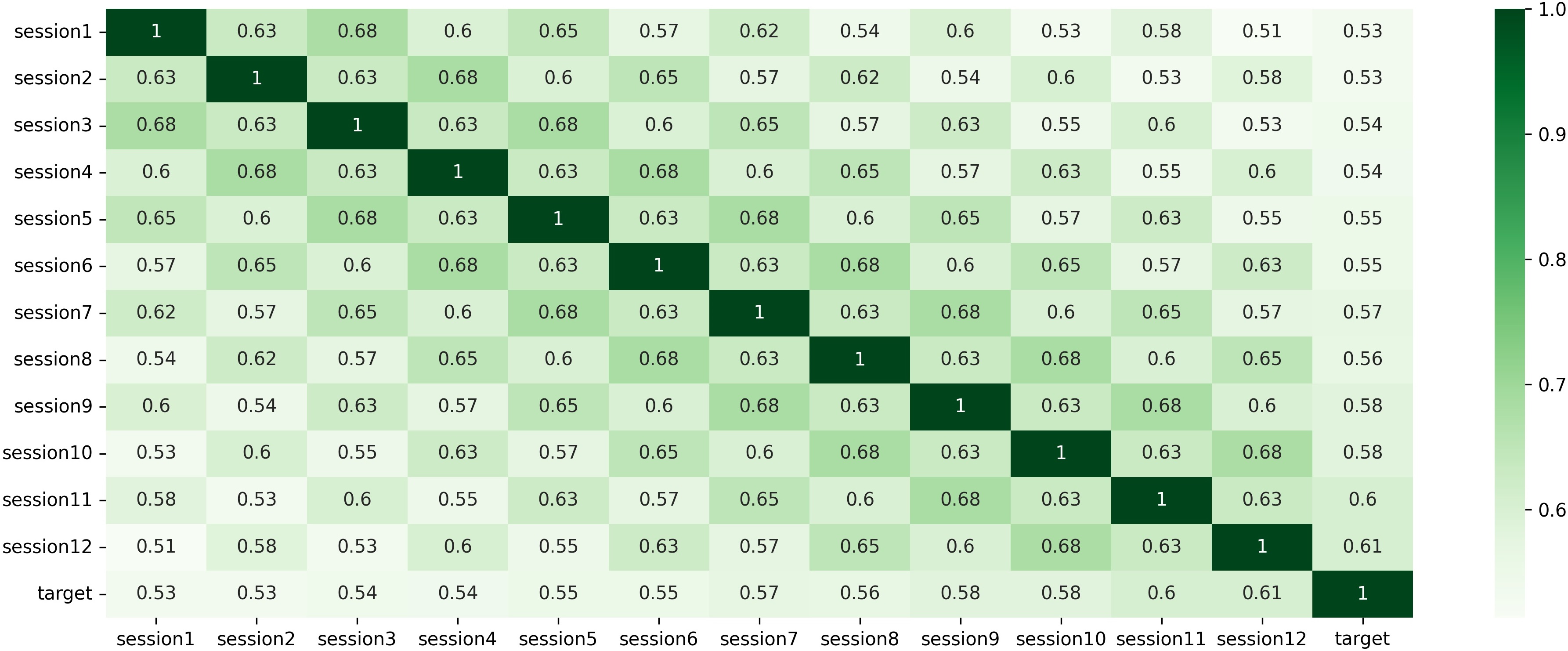}
\caption{Correlation between sessions and the target adherence.}
\label{fig:cor}
\end{figure*}

From the acquisitions collection, we observed that users tend to present high adherence when the previous three to four scheduled acquisitions contain at least one completion. We calculated the Pearson's pairwise correlation per session to verify this observation. Figure \ref{fig:cor} shows that there is a higher correlation between successive sessions and between the last session $S_{12}$ and the target adherence $A$.

\paragraph{\textbf{Static Features.}}
After collecting acquisitions, we processed the users' responses to questionnaires. It is observed that the questionnaires contain multiple null values. Table \ref{tab:nan} shows the percentage of null values per table feature in the provided questionnaires after data cleansing. On the SPQ questionnaire, there are no null values at instance 1 (at the beginning of the experiment). At instance 3, 21.17\% of the users did not respond to questions 1 through 5 and 21.6\% did not respond to question 6. This may be due to the fact that some users have already dropped out at the end of the experiment or some users were not available. Most of the users did not respond to either questionnaire instance on UCLA. More precisely, 53.35\% and 81.21\% of users did not respond at instances 1 and 3, respectively. A similar behavior, i.e., more users have responded at instance 1, is observed on the EQ5D3L questionnaire; 40.82\% and 55.72\% of users did not respond at the beginning and the end of the experiment, respectively. Finally, for the UTAUT, which is only administered at instance 3, we observed that 25.27\% of users did not respond. Based on the response rates at the end of the experiment, approximately 25\% of the initial users dropped off or were unavailable. The rest 80\% responded to the questionnaires, which indicates that most users used the application.

\begin{table*}[htb!]
  \centering
  \caption{Null values per table feature in the questionnaires.}
  \label{tab:nan}
  \begin{tabular}{ lccc }
    \toprule
    \textbf{Table} & \textbf{\ Feature\ } & \textbf{\ Instance\ } &   \textbf{Null Values (\%)}\\
    \midrule
    SPQ &   {Q1, Q3, Q5}  & 1 &   0.00\\
    SPQ & {Q2, Q4, Q6} & 1 & 40.60\\
    SPQ &   {Q1-Q5} & 3   &   21.17\\
    SPQ &   {Q6}    & 3   &   21.60\\
    UCLA & {all}  & 1  &   53.35\\
    UCLA & {all}    & 3  &   81.21\\
    EQ5D3L & {all}   &  {1} &   40.82\\
    EQ5D3L& {all} &   {3} &   55.72\\
    UTAUT& {all} &    {3} & 25.27\\
    \bottomrule
  \end{tabular}
\end{table*}

To assess whether the static responses of questionnaires provide useful features for a classification algorithm, we calculated the Cronbach's alpha \cite{cronbach_coefficient_1951}. The Cronbach coefficient measures the internal consistency of a survey, i.e., it is a reliability indicator. In other words, it measures whether the provided answers in the questionnaires are closely related. Table \ref{tab:cronbach} reports the calculated $\alpha$ of the questionnaires. The higher the value, the more reliable the questionnaire is. From the calculated alphas, the SPQ, UTAUT and EQ5D3L questionnaires have alpha values $> 0.5$. The higher values of alpha are consistent with the lower null rates observed in Table \ref{tab:nan}. Therefore, we are more likely to end up with low prediction rates using only the static features collected from the questionnaires, as they contain multiple null values, which are reflected in the Cronbach's alpha.

\begin{table*}[htb!]
  \centering
  \caption{Cronbach's alpha of the questionnaires.}
  \label{tab:cronbach}  
  \begin{tabular}{ lcc }
    \toprule
    \textbf{Questionnaire} & \textbf{\ Instance\ }  & \textbf{Alpha} \\
    \midrule
    SPQ & 1   &  0.7596   \\
    SPQ & 3    & 0.7587    \\
    UCLA & 1   &   0.3807  \\
    UCLA & 3   &   0.1460   \\
    EQ5D3L & 1 &   0.6624   \\
    EQ5D3L & 3 &   0.6679    \\
    UTAUT & 3  &   0.8481  \\
    \bottomrule
  \end{tabular}
\end{table*}

In addition to the questionnaires, the demographic features of each user are included in the dataset. Table \ref{tab:demographic} reports the minimum, maximum, mean and mode characteristics of the participants. The mode entry concerns the most frequent occurrence in the dataset. From the socio-demographic characteristics, it is observed that most users are elderly people with limited technological level, which is expected since the target group of such apps is elderly people. 

\begin{table*}[htb!]
  \centering
  \caption{Demographic characteristics of the participants.}
  \label{tab:demographic}  
  \begin{tabular}{ lccccc }
    \toprule
    \textbf{Characteristic}    &   \textbf{Min\ } & \textbf{Max} & \textbf{Mean}   & \textbf{Mode}  \\
    \midrule
    \textbf{Year}   &   1924   &   1974    &   1944.81  & 1943\\
    \textbf{Education} &    0   &   8   & 3.25    & 1\\
    \textbf{Technology} &   1   &   3   &  1.68  & 1\\ 
    \textbf{Living environment}    &   1   &   2   & 1.12    & 1\\
    \textbf{Living conditions} &   1   &   2   & 1.12  &    1\\
    \textbf{Living status}  &   1   &   2   & 1.67  &   2\\
    \textbf{Use case}   &   3   &   7   & 5.53  & 6\\
    \bottomrule
  \end{tabular}
\end{table*}

After collecting and joining the dynamic session features and the static features of users, we end up with seven different versions of the dataset. We did not include the logging information, as preliminary local experiments showed no improvement in the classifiers. We plan to explore the influence of logging information in the future.

Each generated dataset is an enhancement of the previous one, i.e., the features are gradually increased. We start with Dataset 0, which contains only the acquisitions of users. At each step, a join operation is performed between the features of the previous dataset and the additional columns. We end up with Dataset 6, which contains the acquisitions and any static information. Table \ref{tab:datasets} summarizes the generated datasets. Note that for the SPQ, EQ5D3L and UCLA questionnaires, we have included the responses from both instances 1 and 3, while the UTAUT questionnaire concerns the instance 3 of the experimental period.

\begin{table*}[htb!]
  \centering
  \caption{The generated datasets.}
  \label{tab:datasets}  
  \begin{tabular}{ lccc }
  \toprule
    \textbf{Dataset Name}    &   \textbf{Features} & \textbf{\#Columns}  \\
    \midrule
    \textbf{Dataset 0}   &   Acquisitions   &   12 \\
    \textbf{Dataset 1} &    Dataset 0 + Timestamp   &   15   \\
    \textbf{Dataset 2} &    Dataset 1 + Demographic  &   22  \\
    \textbf{Dataset 3}    &   Dataset 2 + SPQ   &   34 \\
    \textbf{Dataset 4} &   Dataset 3 + UCLA   &  74  \\
    \textbf{Dataset 5}  &   Dataset 4 + EQ5D3L   &   84   \\
    \textbf{Dataset 6}    &   Dataset 5 + UTAUT   &   115  \\
    \bottomrule
  \end{tabular}
\end{table*}

\paragraph{\textbf{Final Datasets and Imbalance.}}
Each generated dataset contains the features of 463 users. There are 84111 instances, i.e., session tuples from August 2018 to March 2021, with 75.09\% (63159 instances) of them presenting low adherence. Note that these statistics relate to the challenge's combined datasets from Phases I and II. Hence, we end up with an imbalanced dataset, as most samples present low adherence. This is also observed previously from the distribution of average acquisitions in Figure \ref{fig:avg}.

Since the target adherence is highly imbalanced, we employ four oversampling techniques to reduce skewness. First, we performed random oversampling, i.e., we randomly duplicated the samples belonging to the high-adherence class to achieve balance. Second, we used the SMOTE algorithm \cite{chawla_2002_smote}, which is an oversampling technique that generates synthetic samples using the k-nearest neighbors of the samples belonging to the minority class. Third, we employed the ADASYN algorithm \cite{He2008Adasyn}, which generates synthetic data based on the samples of the minority class, which are harder to learn. Finally, we used the CTGAN method \cite{xu2019modeling} to generate synthetic data. The CTGAN is a Generative Adversarial Network (GAN) approach that learns the data distribution from a given dataset and generates synthetic samples using deep learning. After data generation using CTGAN we selected the samples belonging to the minority class and expanded the original feature set. The results with and without oversampling are given in Section \ref{evaluation}.

\paragraph{\textbf{Handling Null Values and Normalization.}}
Machine learning models are trained by exploiting numeric data and therefore, incomplete features should be either removed or transformed. To handle the missing (null) values, we performed mode imputation, i.e., we selected the most frequent observation of a feature and transformed the null values.   

Since the feature ranges are diverse, e.g., the education level $\in \{1, \dots, 8\}$ and the technological level $\in \{1, 2, 3\}$, we employed a normalization step. The normalization step improves the performance of a machine learning algorithm and prevents the model from being biased towards features with higher values. We used the MinMax Scaler to transform the static features to the range $[0, 1]$. 

\subsection{The Issue of Duplicate Session Tuples}
Before describing our runs and results, we raise the issue of duplicate session data. Possibly, the most important feature items of the MAHA dataset are the acquisition data, organized in 12-tuple sessions. However, assuming only the values low and high for each acquisition period, a 12-tuple session can have at most $2^{12} = 4096$ distinct values, whereas the Dataset 0 contains 3948 distinct rows out of 84111. 

This is the case even if we consider the number of acquisitions per session. The highest number of acquisitions observed in each session is 4, i.e., $S_i \in \{0, 1, 2, 3, 4\}$. The possible number of distinct values is $5^{12}$, whereas the dataset has only 26924 unique 12-tuple sessions. Most samples are included multiple times, even considering the number of acquisitions per session, while the corresponding target adherence for a duplicate session can be zero or one.  More precisely, Dataset 0 contains 59663 entries with at least two equal observations. In this dataset, we observed a session tuple included 30892 times. From the 30892 duplicate entries, 30162 have low and 730 have high target adherence, respectively. Consequently, the feature space in the dataset is highly skewed, which may prevent a classification algorithm from correctly distinguishing a high-adherence sample. 

\begin{figure*}[htb!]
    \centering
    \begin{subfigure}[b]{0.48\textwidth}
        \centering
        \includegraphics[width=\textwidth]{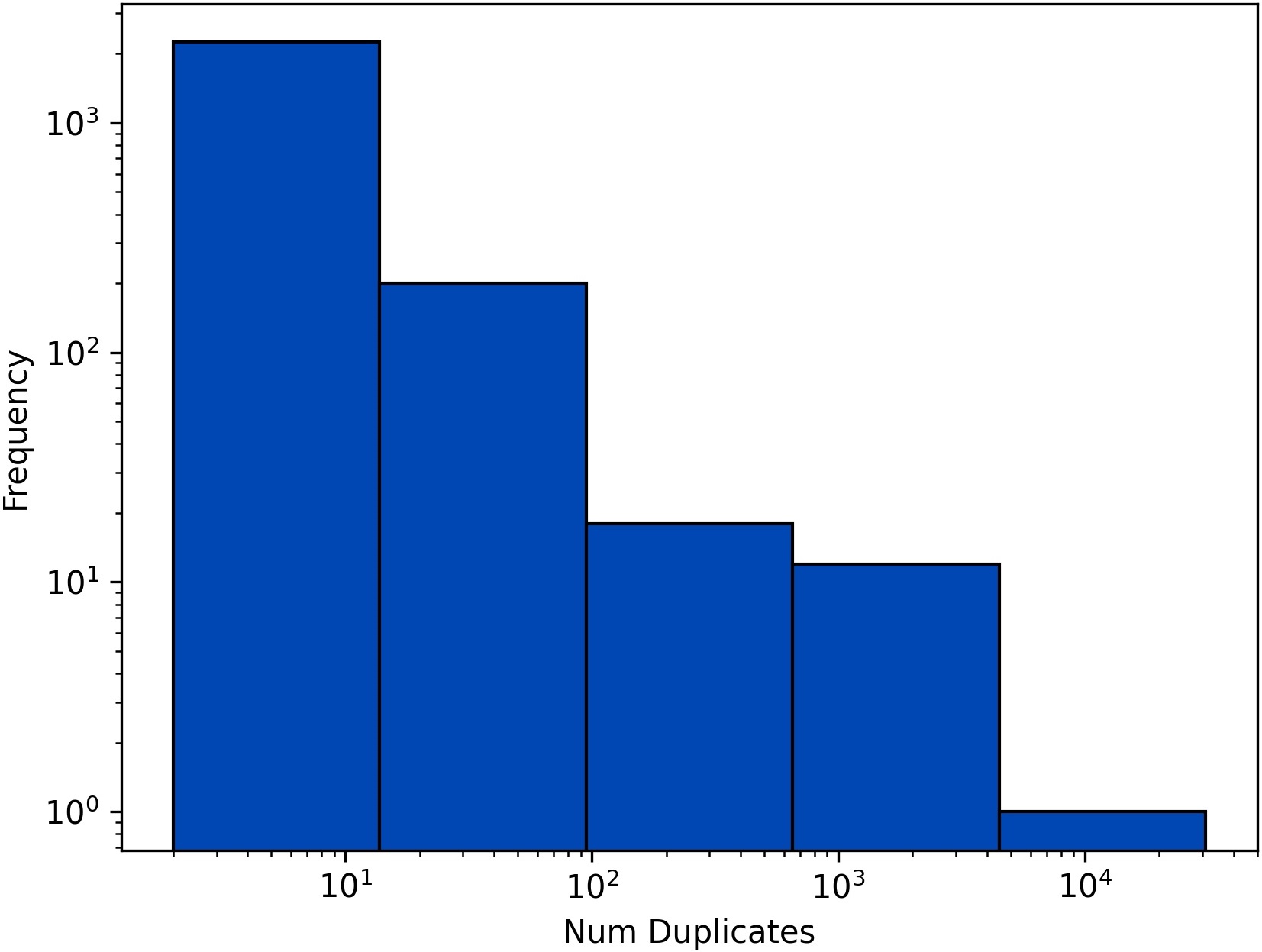}
        \caption{Dataset 0.}
    \end{subfigure}
    \begin{subfigure}[b]{0.48\textwidth}
        \centering
        \includegraphics[width=\textwidth]{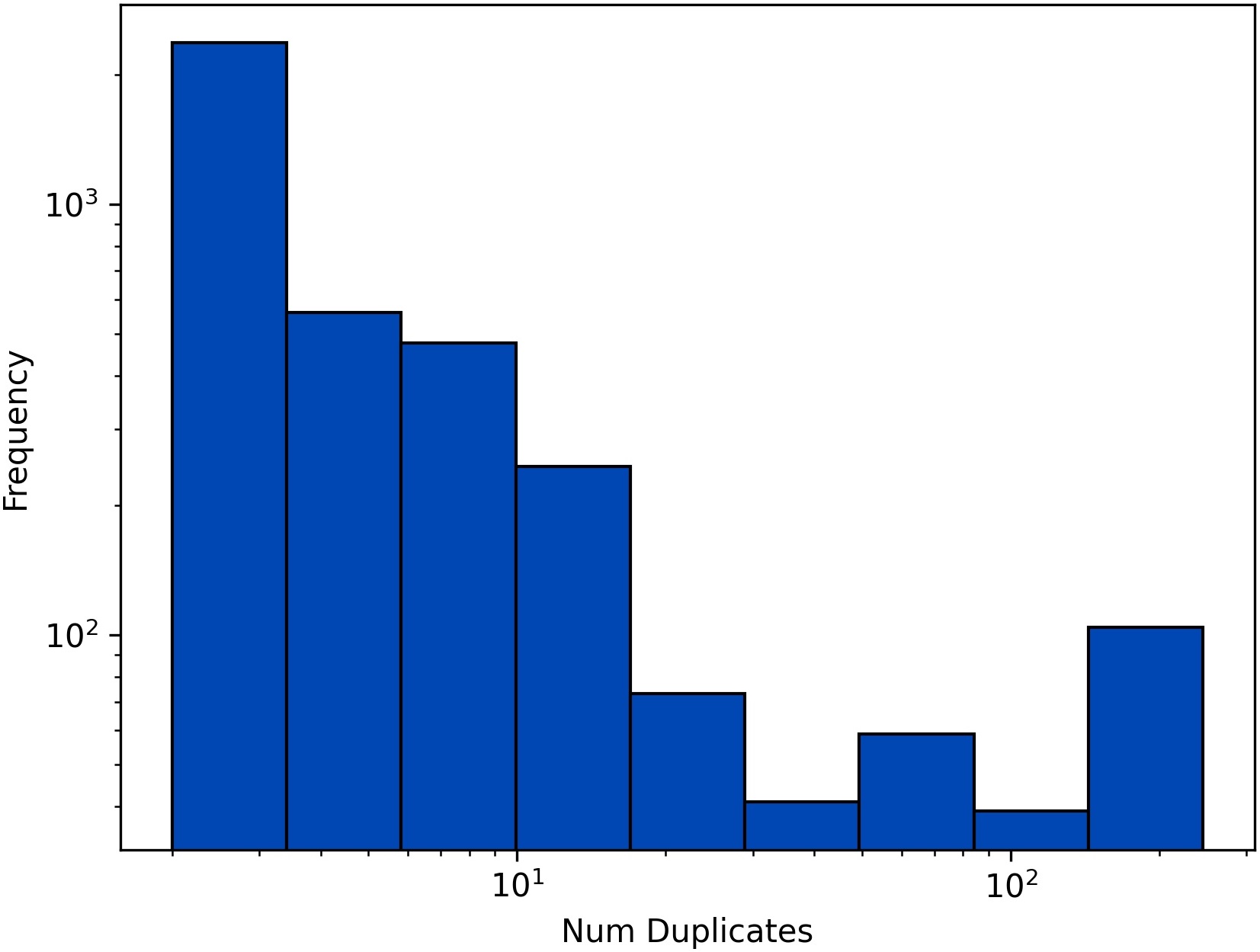}
        \caption{Dataset 1.}
    \end{subfigure}%
    ~~
    \begin{subfigure}[b]{0.48\textwidth}
        \centering
        \includegraphics[width=\textwidth]{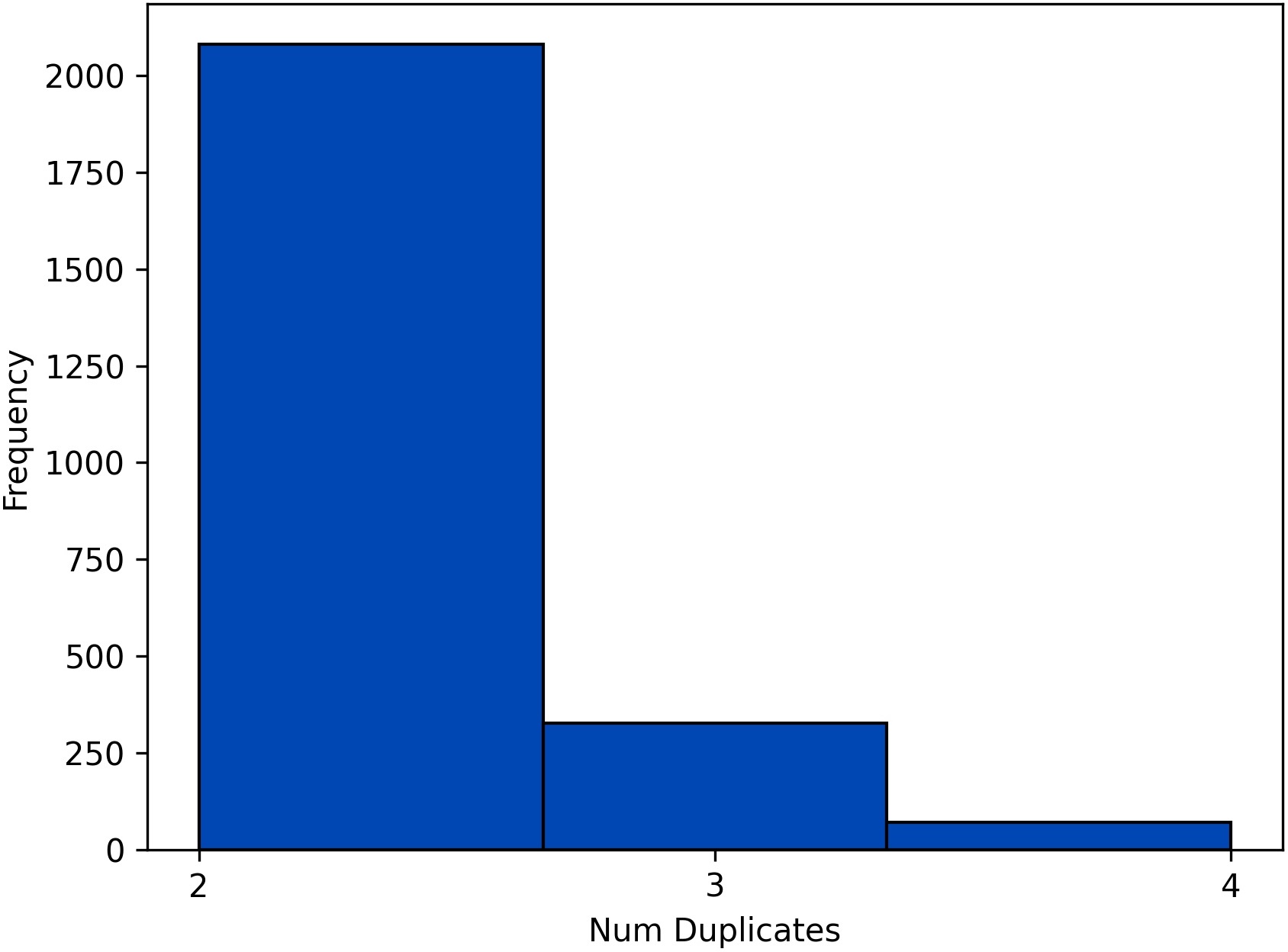}
        \caption{Dataset 2.}
    \end{subfigure}
\caption{Distribution of duplicates in Datasets 0, 1 and 2 using the number of acquisitions per session representation.}
\label{fig:duplicates}
\end{figure*}

Similarly, Dataset 1 contains 46390 duplicate entries and 15 session tuples with more than 220 duplicates. Finally, Dataset 3 contains 5422 duplicates and 70 session tuples with 4 observations. From Dataset 3 onwards, the issue of duplicate data is eliminated using the number of acquisitions representation. Figure \ref{fig:duplicates} shows the distribution of duplicate entries. The X-Axis shows the number of duplicate entries and the Y-Axis the corresponding frequency of observations.

The issue of duplicate data raises concerns regarding the separability among samples for the first three datasets, as a machine learning algorithm will be biased towards the most observed target value for the duplicate sessions. This can lead to the concept-learning problem \cite{Jo2004Dis}, making it difficult for machine learning models to classify the instances of the minority class correctly. Hence, at the inference stage, the model is very likely to predict the most frequent target value observed at the training stage. Since most observations relate to low adherence, the model will be biased towards this class and generalization is challenging to achieve.

\section{Runs and Results}
\label{evaluation}
In this section, we describe our experiments and report the results of the machine learning algorithms used for evaluating our classifiers on the generated datasets.
\subsection{Local Evaluation}

We used several classification algorithms, including Random Forest (RF), k-NN, XGBoost \cite{xgboost} and a Multi-Layer Perceptron (MLP). Conventional machine learning algorithms are implemented on \textit{scikit-learn} \cite{pedregosa2011scikit}, the XGBoost classifier using the official python library\footnote{\url{https://xgboost.readthedocs.io/en/stable/}} and the MLP is implemented on PyTorch \cite{Pytorch}. The programming language is Python and the operating system is Ubuntu 20.04. 

Each classifier is evaluated locally on each of the seven datasets. The number of estimators to the RF classifier is 200 and the $k$ in k-NN is 30. For the XGBoost classifier, we set the maximum tree depth to 10, the $l2$ regularization to 1 and the minimum child weight to 0. The architecture of the MLP is $\{1024, 512, 256, 128\}$, the batch size is set to 128, the learning rate is 0.001, the optimizer is Adam, and the loss function is the cross-entropy loss. The maximum number of training epochs is set to 50 and an early-stopping callback is implemented to avoid over-fitting. The parameters for the classifiers are set after a small grid search.

For evaluating the classifiers, we use 10-fold cross-validation and the performance metrics are \textit{Accuracy}, \textit{Sensitivity} and \textit{Specificity}. The total \textit{score} is calculated using the geometric mean of sensitivity and specificity, i.e., $\text{score} = \sqrt{\text{Sensitivity} \cdot \text{Specificity}}$.

\begin{table*}[htb!]
  \centering
  \caption{Results with 10-fold cross-validation using the Dataset 3.}
  \label{tab:results}  
  \begin{tabular}{ lcccccc }
    \toprule
    \textbf{Classifier} & \textbf{\ Accuracy\ } & \textbf{\ Specificity\ } &  \textbf{\ Sensitivity\ } & \textbf{Score}\\
    \midrule
    \textbf{RF}  &  0.8934   &   0.9468  &   0.7322 &   0.8326  \\
    \textbf{k-NN} ($k=30$)  &   0.8889   &   0.9589  &   0.6778  & 0.8061\\
    \textbf{XGBoost} &  0.8966  &  0.9438  &  0.7727  &  0.8540\\
    \textbf{MLP} & 0.8940  & 0.9596 & 0.7671 &   0.8451\\
    \bottomrule
  \end{tabular}
\end{table*}

The local evaluation shows that the highest score is generated using Dataset 3, which includes the session instances, demographic features and SPQ questionnaire answers. The average testing results after 10-fold cross-validation for the considered classifiers using Dataset 3 are given in Table \ref{tab:results}. The evaluation scores using the rest of the datasets will be included in the long version of this work. 

From the results, the MLP and the XGBoost models provide higher classification performance. More precisely, these models are more likely to correctly predict high adherence, which is reflected in the sensitivity measure. This is an indicator that MLP and XGBoost can handle some cases, which concern sessions with high adherence, which are not obvious. Nevertheless, conventional machine learning algorithms achieve satisfactory performance and offer the advantage of training speed. Note that all classifiers outperform the majority baseline accuracy (0.7509), i.e. when predicting low adherence for each instance.

\begin{figure*}[htb!]
    \centering
    \begin{subfigure}[b]{0.49\textwidth}
        \centering
        \includegraphics[width=\textwidth]{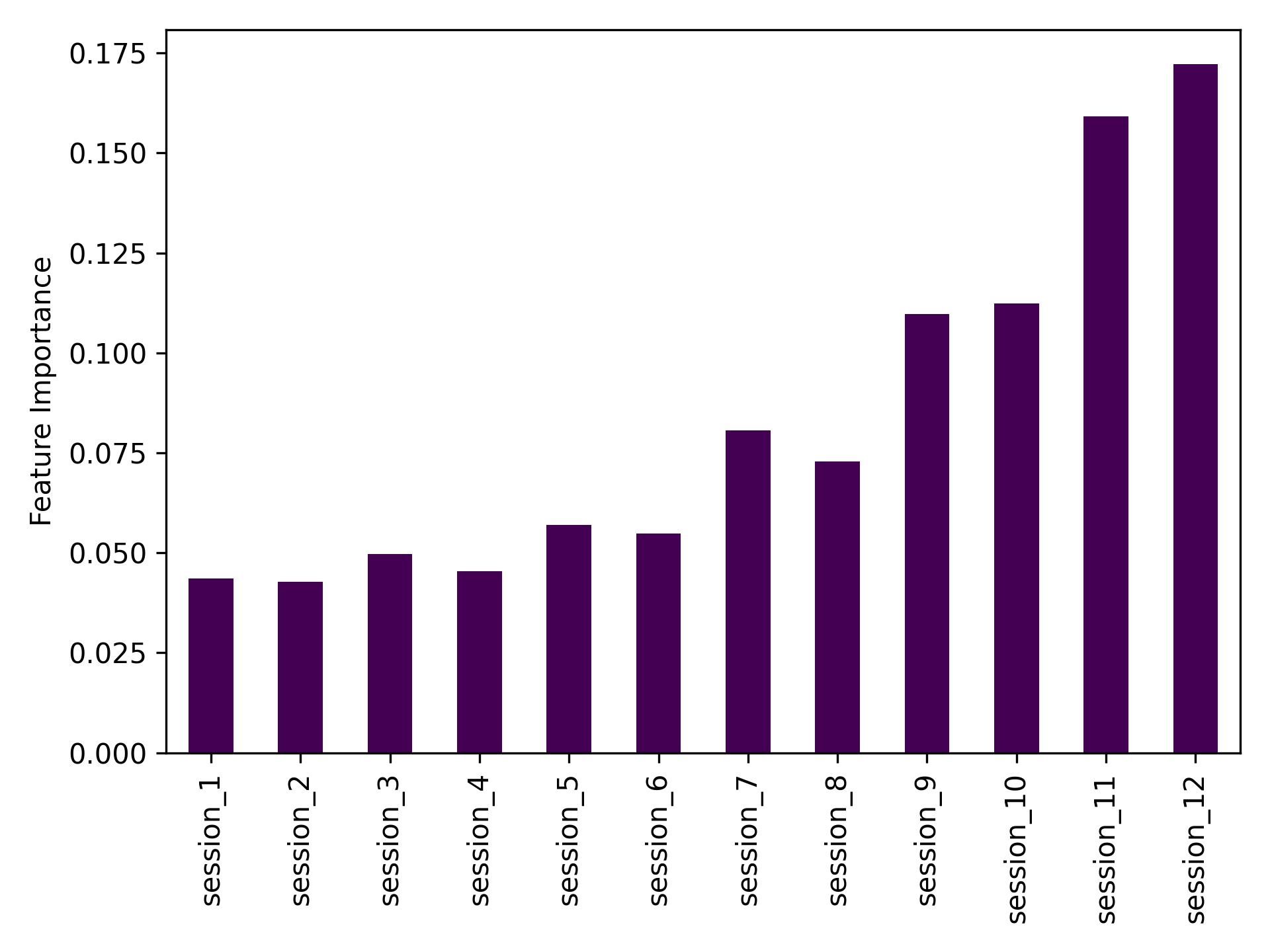}
        \caption{Feature importance on Dataset 0.}
    \end{subfigure}
    \begin{subfigure}[b]{0.49\textwidth}
        \centering
        \includegraphics[width=\textwidth]{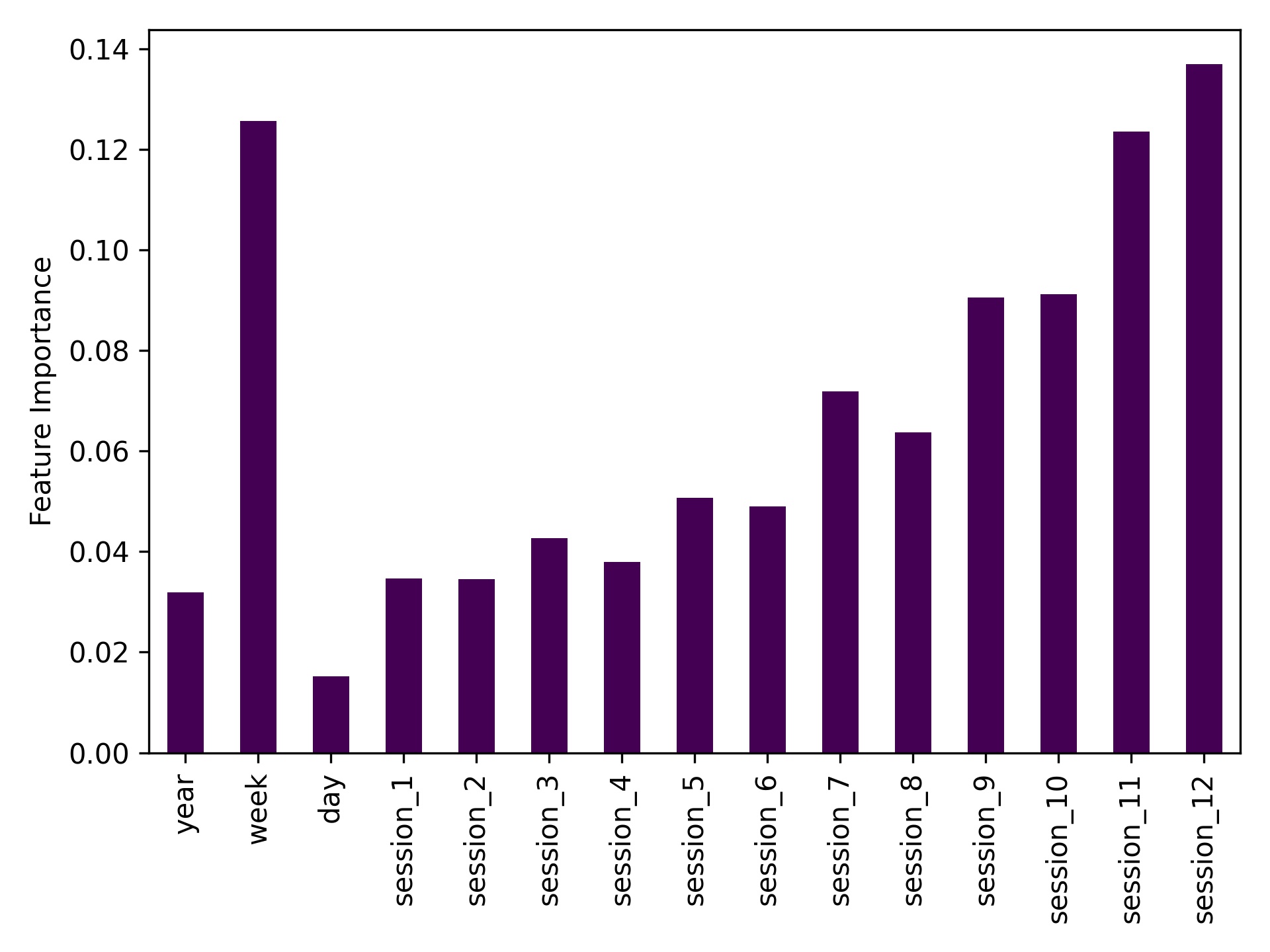}
        \caption{Feature importance on Dataset 1.}
    \end{subfigure}
    \begin{subfigure}[b]{0.49\textwidth}
        \centering
        \includegraphics[width=\textwidth]{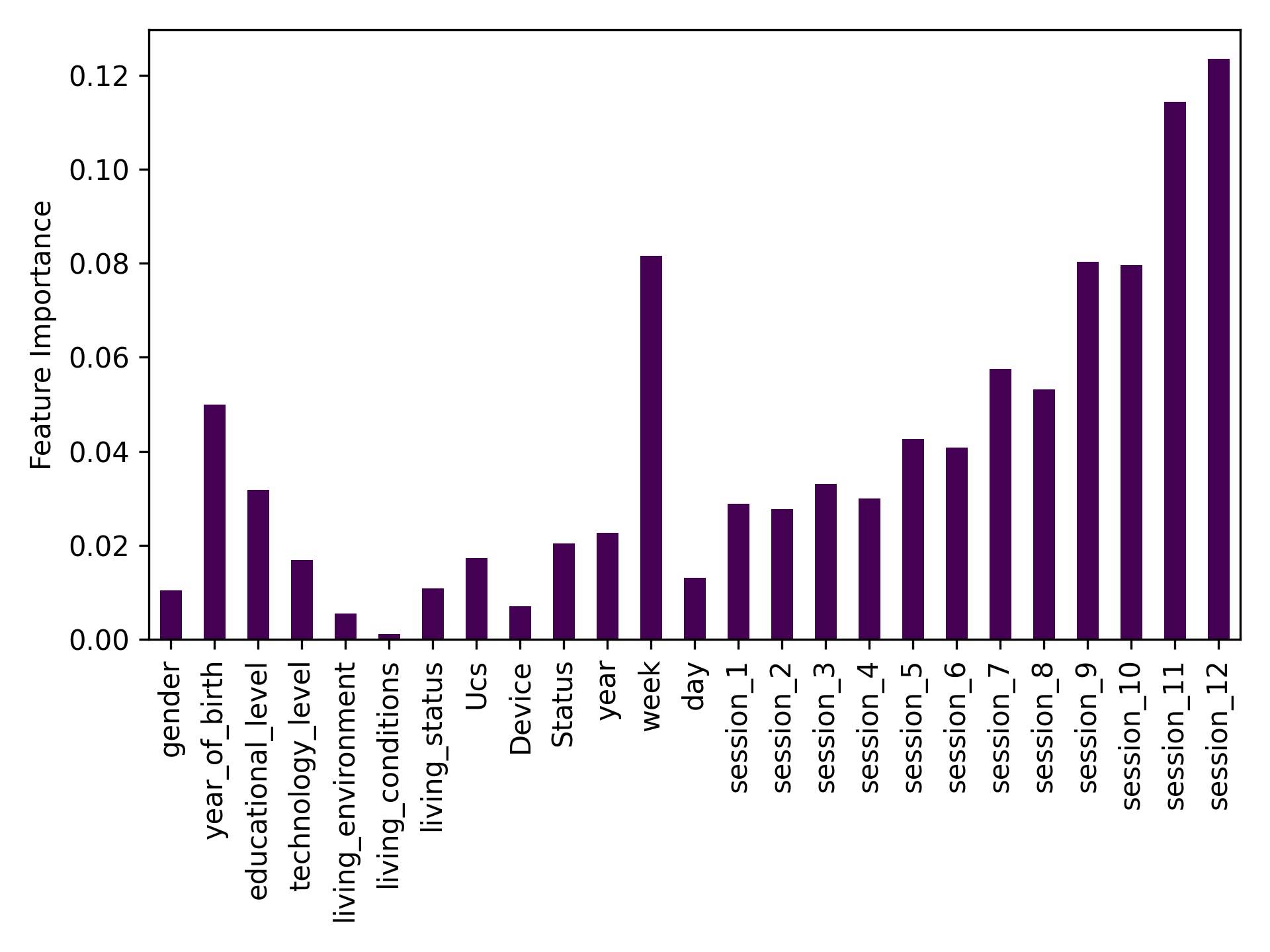}
        \caption{Feature importance on Dataset 2.}
    \end{subfigure}
    \begin{subfigure}[b]{0.49\textwidth}
        \centering
        \includegraphics[width=\textwidth]{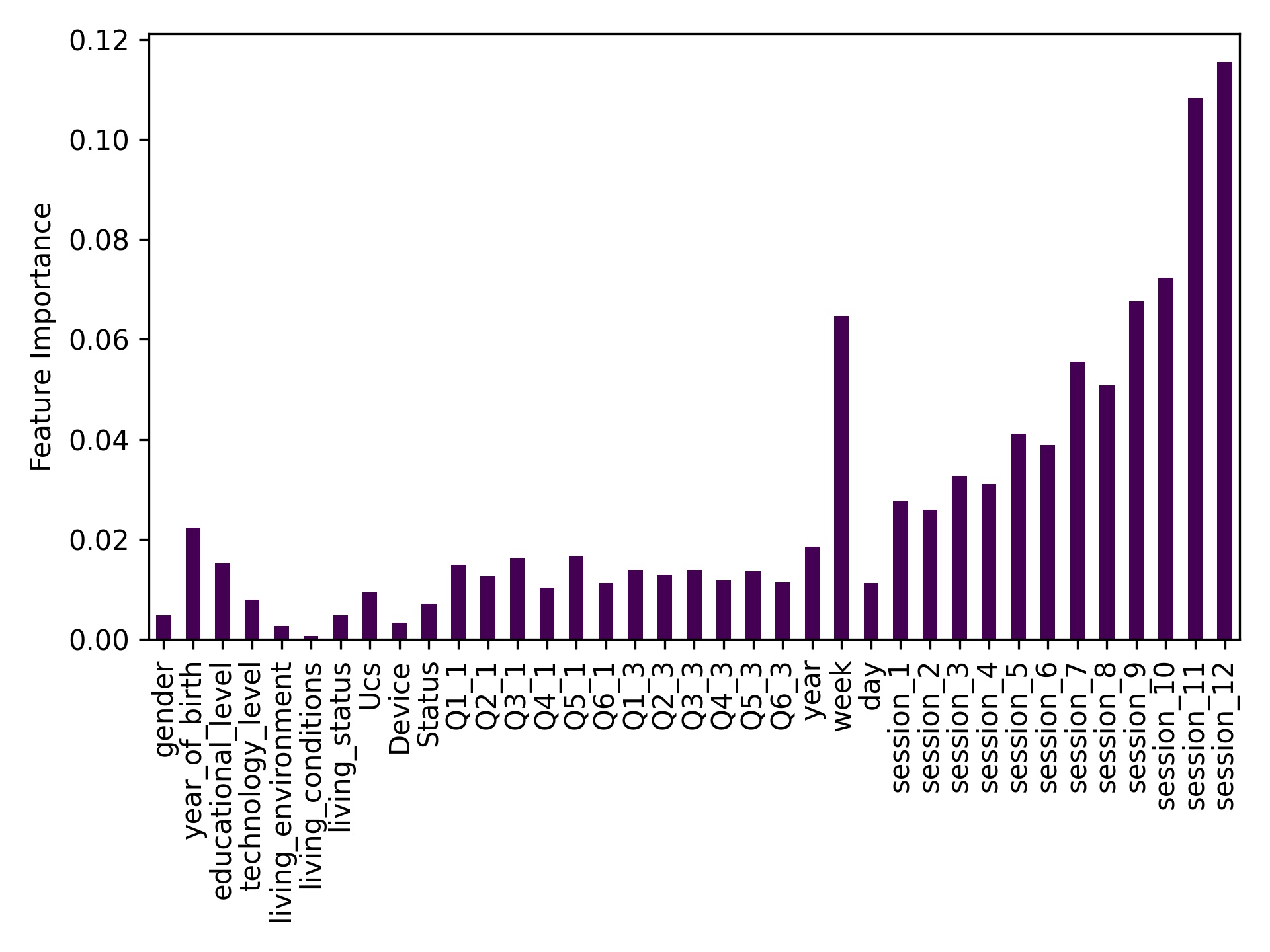}
        \caption{Feature importance on Dataset 3.}
    \end{subfigure}
\caption{Feature importance using a Random Forest classifier with 200 estimators.}
\label{fig:importance}
\end{figure*}

To explain the models' behavior from the input data to the adherence prediction, we calculated the importance of each feature on the RF model. We did not include the feature importance scores for Dataset 4 onwards, since the performance on each classifier decreases. The decreasing performance using more data is attributed to the fact that the questionnaires contain multiple null values and introduce noise to the classification algorithms. Noisy data negatively influence the classification performance on the minority class and thus, lead to poorer overall performance. Figure \ref{fig:importance} shows that the last two acquisition sessions strongly influence the classifier. This behavior is similar on Datasets 0 through 3 regarding the acquisitions' influence to the target adherence and complies with the observation from Figure \ref{fig:cor}. Starting with Dataset 1, the week's number has a decisive role in predicting adherence, as the corresponding importance is almost equivalent to the last two historical acquisitions. This is attributed to specific dates, which may involve holidays. For instance, national-level holidays can influence the usage of an application. In the rest of the datasets, the most influential features are the year of birth and the technological level of users. Intuitively, elderly people with lower technological levels have a higher probability of dropping out, which is reflected in the classifier's importance. 

We employed four oversampling techniques to minimize the impact of the concept-learning problem and reduce skewness on the two classes. More precisely, a random oversampling and generation of synthetic data using the SMOTE, the ADASYN and the CTGAN algorithms are implemented (as mentioned in Section \ref{subsec:preprocessing}). The SMOTE and the ADASYN methods are implemented using the \textit{imbalanced-learn} library \cite{imbalanced-learn}. The CTGAN approach is implemented using the official CTGAN library\footnote{\url{https://github.com/sdv-dev/CTGAN}}. These algorithms generate synthetic data for the minority class, i.e. samples corresponding to high adherence to achieve balance.

Table \ref{tab:oversamplingresults} shows the average classification performance using 10-fold cross-validation of the MLP model using the oversampling methods on Datasets 0 and 3. With oversampling, the skewness problem is minimized, as the classifier can predict the samples belonging to the minority class with a higher probability compared to the baseline evaluation (Table \ref{tab:results}). Each of the four oversampling strategies shifts the classifier's learning towards the minority class and can provide higher scores. 

Based on these results, each oversampling method provides almost equivalent scores and overall, the classifier is not biased towards a class. On Dataset 0, the ADASYN algorithm did not improve the classification performance. This behavior is attributed to duplicate overlapping data between classes and the functionality of ADASYN; it generates synthetic data based on hard-learned samples. Hence, some high-adherence samples that overlap with the majority class are used to generate the synthetic data. However, these data may concern outliers and amplifying them leads to performance degradation. The random oversampling technique slightly outperforms SMOTE, ADASYN and CTGAN algorithms on both datasets 0 and 3. In the future, we will focus on the explainability of the improvement using additional methods.

\begin{table*}[htb!]
  \centering
  \caption{Results with 10-fold cross-validation on the MLP model using oversampling methods.}
  \label{tab:oversamplingresults}
  \begin{tabular}{ lccccccc }
    \toprule
    \textbf{Method} & \textbf{\ Dataset\ } & \textbf{\ Accuracy\ }    & \textbf{\ Specificity\ } &  \textbf{\ Sensitivity\ } & \textbf{Score}\\
    \midrule
    \textbf{Random}  &  Dataset 0   &   0.8661  &   0.8662 &   0.8513 & 0.8811  \\
    \textbf{Random}  &  Dataset 3   &   0.9034  &   0.9098 &   0.9337 & 0.9033  \\
    \textbf{SMOTE}  &  Dataset 0   &   0.8573  &   0.8573 &   0.8611 & 0.8535  \\
    \textbf{SMOTE}  &  Dataset 3   &   0.9001  &   0.8946 &   0.9056 & 0.8998  \\
    \textbf{ADASYN}  &  Dataset 0   &   0.7929  &   0.8215 &   0.7644 & 0.7924 \\
    \textbf{ADASYN}  &  Dataset 3   &   0.8929  &   0.8998 &   0.9426 & 0.8909  \\
    \textbf{CTGAN} & Dataset 0  &   0.8852 & 0.9251 &   0.8032 & 0.8620\\
    \textbf{CTGAN} & Dataset 3 & 0.9004 & 0.8995    & 0.9147    &  0.9071 \\
    \bottomrule
  \end{tabular}
\end{table*}

\subsection{Official Runs at the IFMBE Challenge}
In Phase I of the challenge, we submitted one classifier using the corresponding data, and in Phase II, 10 classifiers using the combined data from Phases I and II. The scores of our submissions are given in Table \ref{tab:official_results}.

\begin{table*}[htb!]
  \centering
  \caption{Official results of the IFMBE Scientific Challenge 2022 (column \textit{Score}) compared to our local evaluation (column \textit{Local Score}).}
  \label{tab:official_results}
  \begin{tabular}{ cccccc }
    \toprule
    \textbf{Phase-Run} & \textbf{Classification Model} & \textbf{\ \ Dataset\ \ }    & \textbf{\ Local Score\ } & \textbf{Score}\\
    \midrule
    I-1 & MLP & Dataset 0  & 0.8376  &  0.7532 \\
    II-1 & Ensemble & Dataset 0    & 0.8436 & 0.7540 \\
    II-2 & MLP & Dataset 0  & 0.8263  & 0.7655 \\
    II-3 & MLP & Dataset 3    & 0.8451 & 0.7356 \\
    II-4 & XGBoost & Dataset 3  &   0.8540 & 0.6824 \\
    II-5 & MLP+Random Oversampling & Dataset 3 & 0.9033 & 0.7031 \\
    II-6 & MLP+SMOTE   & Dataset 3  & 0.8998 & 0.5014 \\
    II-7 & MLP+SMOTE   & Dataset 0  & 0.8535 & \textbf{0.8635} \\
    II-8 & MLP+ADASYN  & Dataset 3  & 0.8909 & 0.5663 \\
    II-9    &   MLP+CTGAN   & Dataset 0 & 0.8620 & 0.7512 \\
    II-10   &   MLP+ADASYN  & Dataset 0 & 0.7924 & 0.8545 \\
    \bottomrule
  \end{tabular}
\end{table*}

We submitted three MLP models, an XGBoost and a voting classification approach for the models trained without oversampling techniques and 6 classifiers using oversampling techniques with the MLP model. More precisely, we generated an ensemble consisting of 8 different classifiers (II-1). The classifiers in the ensemble are: Decision Tree, Random Forest, ExtraTree, Linear SVM, GradientBoosting, Adaboost, Gaussian Naive Bayes and XGBoost. Each classifier predicts the target adherence for a sample and the final classification is generated using the mean of the predicted probabilities. The ensemble method did not outperform the single MLP classifier (II-2), while the difference on the trained MLP model between Phases I (I-1) and II using the same training process is 1\%. 

The first three runs are generated using Dataset 0. In our third and fourth submissions for Phase II (II-3 and II-4), we submitted a MLP and a XGBoost classifier using Dataset 3. The MLP and XGBoost classifiers scores are 0.7356 and 0.6824, respectively. Therefore, the MLP model can better generalize than XGBoost, although the latter performs slightly better in local evaluation.

Finally, we submitted 6 MLP classifiers, trained with the considered oversampling techniques. The oversampling methods on Dataset 3 decrease the classification performance. More precisely, the trained classifiers with the SMOTE and ADASYN algorithms provide a poor classification score of 0.5014 and 0.5663, respectively (II-6 and II-8). The trained model using the random oversampling method (II-5) decreases the classification performance to $\approx 3.5\%$ compared to the submission of the same model using the original dataset (II-3). The degradation is attributed to different distributions of training and validation sets regarding the relation of the users' demographic features to their corresponding acquisition patterns. The last two submissions (II-9 and II-10) concern the ADASYN and CTGAN synthetic data generation algorithms. The generation of synthetic data using CTGAN did not improve the classification performance. The ADASYN method, similar to SMOTE (II-7), improved the classification performance by $\approx 10\%$ compared to the training procedure using the original dataset. Hence, the popular oversampling techniques, i.e., the SMOTE and ADASYN algorithms, led to a remarkable classification improvement. The highest quality model is generated with the SMOTE algorithm, which slightly outperforms ($\approx 1\%$) the corresponding model trained with the synthetic data generated by ADASYN. Our team won first place in the IFMBE Scientific Challenge 2022.

The difference between the local evaluation and the official results on Dataset 0 is $\approx 10\%$ regarding the classification score with the original dataset. Meanwhile, the SMOTE oversampling algorithm (II-8) provides almost equivalent scores using Dataset 0 on the local and official evaluation. This oversampling technique provided the highest score on the official results. On the other hand, there is a huge classification performance degradation on Dataset 3 using either the original datasets or the oversampling techniques compared to the local evaluation. This behavior of our classifiers is attributed to dataset shift, i.e., the training and validation sets may follow different distributions. We plan to explore in more detail the differences in the classification performance when the validation set is made available by the organizers of the challenge.

\section{Conclusion}
\label{conclusion}
Providing better solutions to the adherence problem has social implications on the quality of life of elderly people and ultimately, it can contribute to the improvement of healthy ageing. The corresponding prediction problem defined in the IFMBE Scientific Challenge 2022 was to predict users' adherence in the form of a binary classification task.

We participated in this challenge using pre-processing techniques to generate the final datasets and tried several binary classification methods, including XGBoost, MLP and an ensemble of classifiers in the official runs. The MLP and XGBoost classifiers slightly outperformed conventional machine learning algorithms in the local evaluation, while conventional classifiers have the advantage of faster training. The local evaluation and the official results show that the dataset's features, besides the consecutive session acquisitions, did not positively impact the adherence prediction. The highest classification performance is generated using oversampling, specifically with the SMOTE algorithm.

The most challenging issue during our experimentation was the raw dataset interpretation. In particular, the session interpretation in days and the interpretation of the questionnaires led to the identification of several limitations which possibly had a negative impact on the classification performance. The content-learning problem, the presence of noisy data and, in general, the imbalanced dataset lead to high misclassification rates in the minority class. Thus, we employed oversampling techniques to minimize the impact of the identified limitations. 

In the future, we will focus on eliminating these problems using other techniques and analyze the features that lead to performance degradation when the evaluation set is made available.

%
%
\renewcommand{\bibsection}{\section*{References}}
\bibliographystyle{spbasic}
\bibliography{refs}

\end{document}